\documentclass[conference]{IEEEtran}

\usepackage{cite}

\ifCLASSINFOpdf
\else
\fi

\usepackage{xcolor}
\usepackage{chngcntr}
\usepackage{amsmath}
\usepackage{graphicx}
\usepackage{amsfonts}

\usepackage{chngcntr}
\usepackage{enumitem}
\setlist[itemize]{leftmargin=*}

\begin{document}

\title{Automatic Malware Description via Attribute Tagging and Similarity Embedding}

\author{
    \IEEEauthorblockN{Felipe N. Ducau \IEEEauthorrefmark{1},
                      Ethan M. Rudd \IEEEauthorrefmark{2},
                      Tad M. Heppner \IEEEauthorrefmark{3},
                      Alex Long \IEEEauthorrefmark{3},
                      and
                      Konstantin Berlin \IEEEauthorrefmark{1}}
    \IEEEauthorblockA{\IEEEauthorrefmark{1} Sophos AI, equal contribution.}
    \IEEEauthorblockA{\IEEEauthorrefmark{2} FireEye, research done while at Sophos AI}
    \IEEEauthorblockA{\IEEEauthorrefmark{3} Sophos AI}
}

\maketitle

\begin{abstract}
With the rapid proliferation and increased sophistication of malicious software (malware), detection methods no longer rely only on manually generated signatures but have also incorporated more general approaches like machine learning detection. Although powerful for conviction of malicious artifacts, these methods do not produce any further information about the type of threat that has been detected neither allows for identifying relationships between malware samples.
In this work, we address the information gap between machine learning and signature-based detection methods by learning a representation space for malware samples in which files with similar malicious behaviors appear close to each other. 
We do so by introducing a deep learning based tagging model trained to generate human-interpretable semantic descriptions of malicious software, which, at the same time provides potentially more useful and flexible information than malware family names.

We show that the malware descriptions generated with the proposed approach correctly identify more than 95\% of eleven possible tag descriptions for a given sample, at a deployable false positive rate of 1\% per tag.
Furthermore, we use the learned representation space to introduce a similarity index between malware files, and empirically demonstrate using dynamic traces from files' execution, that is not only more effective at identifying samples from the same families, but also 32 times smaller than those based on raw feature vectors.

\end{abstract}

\IEEEpeerreviewmaketitle

\section{Introduction}
\begin{figure}[!t]
    \centering
    \includegraphics[width=0.9\linewidth]{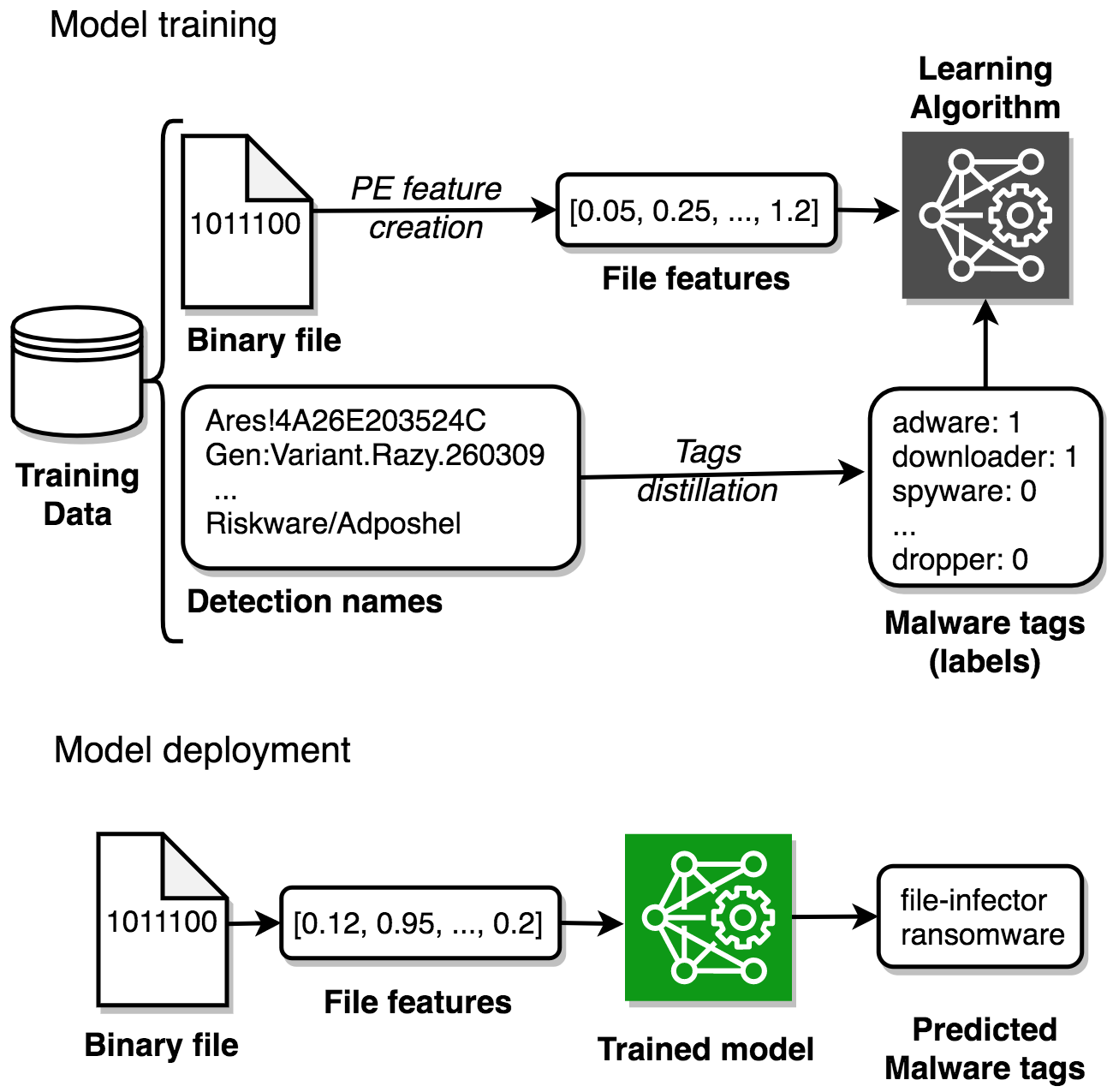}
    \caption{Overview of the proposed system architecture for a malware tagging model. In model training phase (top figure) we use our database of binary files along with their associated detection names to train a machine learning model that learns a non-linear mapping between PE file features and malware tags. At deployment time (bottom figure), the proposed model generates descriptive tags for new, previously unseen, executable files.}
    \label{fig:system}
\end{figure}

Whenever one or more malicious files are found in a computer network, the first step towards remediation is to understand the nature of the attack in progress. Knowing the malicious capabilities associated with each suspicious file gives important context to network defenders which helps them define and prioritize counter-measures. 

Generally,  anti-virus (AV) or anti-malware solutions provide a \emph{detection name} when they alert about potentially harmful files detected in a machine as a way to provide this context. These detection names usually come from specific signatures written by reverse engineers to identify particular threats, therefore encoding expert knowledge about a given malware sample. 
While this is theoretically useful for categorizing known malware variants, differing malware naming conventions among vendors have led to detection names that are inconsistent and highly vendor-specific \cite{inconsistent_naming, finding_inconsistencies_maggi}. For example, \emph{Worm.Ludbaruma.B} and \emph{Win32.Worm.VB.k}, are detection names produced by two different vendors for the same file. The problem of inconsistent naming conventions has been compounded due to more feature-rich malware and increased quantities of threats over time. Moreover, some detection names serve only as unique identifiers and do not provide actionable information about what type of harm the malicious sample could do if it infects a system (e.g. \emph{Gen:Variant.Razy.260309} or \emph{Trojan (005153df1)}). 
 
When a novel malware variant appears, applying existing detection names, or even measuring similarity with known malicious files is problematic, since current rule-based signatures will likely not trigger on these variants at all. 
Machine learning (ML) malware detectors have the potential to identify these new malware samples as malicious, but generally do not provide further information about the type of threat encountered neither on how it relates with the universe of known malware.

In this paper, with existing detection naming issues in mind, we propose to use Semantic Malware Attribute Relevance Tagging (SMART) to approach malicious software description. In contradistinction to prior malware (family) detection names, this semantic malware attribute tags approach yields human interpretable, high level descriptions of the capabilities of a given malware sample. They can convey different types of information such as purpose (`crypto-miner', `dropper'), family (`ransomware'), and file characteristics (`packed'). SMART tags are related to malware family names in the sense that they attempt to describe how a piece of malicious software executes and the intent behind it. However, unlike malware family names, malware tags are non-exclusive, meaning that one malware campaign (or family) can be associated with multiple tags and a given tag can be associated with multiple malware families.

The number of tags is also inherently bounded by types of malicious behavior and chosen granularity in description. Thus, a fixed number of tags can roughly describe all malicious samples, even when the number of malware families increases dramatically. Because of this, the tagging approach makes the task of malware description suitable to be addressed with machine learning methods.

Semantic attribute tags also serve as a common ground to integrate knowledge from multiple sources or detection technologies since they do not presume standard naming conventions. For the experiments in this paper, we derive tags by leveraging the underlying knowledge encoded in detection names from different anti-malware vendors in the industry, although the general framework applies whenever we have multiple analyses of the same file. 

Using our derived tags, we then train a multi-label deep neural network to automatically predict tags for new (unseen) files in real time. Our approach only assumes access to the files' static binary representations. We find that our network yields impressive performance on the tag prediction task. It does so by learning a low dimensional Euclidean representation space in which malware samples with similar characteristics are close to each other. Figure \ref{fig:system} provides an overview of the proposed tagging system: during the model training phase we use binary files and malware descriptions generated by multiple expert systems to train a machine learning model, that can later be deployed to automatically produce descriptive tags for new files in a fraction of a second.  

Moreover, the proposed approach of describing malicious capabilities by learning a low dimensional embedding space (and associated embedding function) for malicious files enables us to compare malware samples by \textit{semantic similarity} in terms of \textit{type} of malicious content. This compressed representation further allows for efficient indexing, searching and querying of malware corpora along explainable dimensions. This ability is particularly useful for identifying new samples of a novel malware campaign from which we only have identified a small number of samples -- even one. As we will show, this similarity metric in latent space also opens the door to novel applications in the context of endpoint detection and response (EDR) technologies, such as natural language queries, mapping how a given (potentially novel) malware campaign relates to and compares with known malware, and alerts prioritization based on malicious content, among others.

The primary contributions of this paper are as follows: 

\begin{enumerate}
    \item We introduce the task of automatic malicious tag prediction for malware description, and propose a Joint Embedding neural network architecture and training methodology that allows us to generate descriptive tags for malware files with high true positive rates at low false positive rates in a static fashion (without executing the software).
    
    \item We empirically demonstrate that our neural networks learn a compact, yet expressive representation space for binary files which is informed by their malicious capabilities.
    
    \item We propose a novel file-file similarity index based on the representation space of our neural networks that enables for efficient indexing and searching in large malware corpora along interpretable dimensions.

\end{enumerate}

The remainder of this paper is structured as follows: In Section \ref{sec:background} we discuss existing approaches to malware description, particularly family names and hierarchies and review attempts to establish industry-wide standard naming conventions. We also discuss related research in the machine learning for information security (ML-Sec) space as well as similar applications in other domains, such as image tagging, music information retrieval, and semantic facial attribute recognition. In Section \ref{sec:mal_tags}, we define the concept of describing malware with semantic tags, and present one method for deriving semantic tags from binary executable files by aggregating detection names from multiple vendors in the industry, that can easily scale to millions of files.
We then formalize the problem of malware characterization as a tagging problem in Section \ref{sec:tagging_with_dnn}, and propose two neural network architectures for tag prediction. In Section \ref{sec:experiments}, we train several neural networks and evaluate their performance on the tagging task. We analyze our results and their ramifications in Section \ref{sec:results}. In Section \ref{sec:discussion} we analyze the latent space learned by the proposed neural networks and identify practical applications in the domain of information security.
Finally, we present conclusions and propose directions for future research in Section \ref{sec:conclusion}.

\section{Background and Related Work}
\label{sec:background}

In this section we revisit the idea of malware description by using family classification, review the concept of attribute tagging in other domains, and survey related machine learning approaches in the field of machine learning for information security (ML-Sec).

\subsection{Malware Family Categorization}
\label{sec:malware_families}
Identifying the family and variant of a particular malicious sample can provide important intelligence to the end user, administrator or security operator of a system about what type of attack might be underway. This extra contextual information can help define a remediation procedure, identify possible root causes, and evaluate the severity and potential consequences of the attack. 
In fact, numerous vendors provide in their websites detailed information about popular family/variant information, with associated description of what that variant does, and suggest how a particular piece of malware can be removed.
Without such identifying information, we are left only with the offending file itself as its own description. Unless some reverse engineering effort is taken, which can be costly, it is difficult to discern much about the internals of the file.

The idea to identify all malware under a consistent family naming scheme across multiple vendors has been around for decades. It first came to the attention of the security community in the early 1990's and prompted the Computer Antivirus Research Organization (CARO) to propose a first naming convention in 1991 \cite{naming91}, which was was later extended to add coverage for new kinds of malicious software such as backdoors, trojans, joke programs and droppers, among others \cite{caro2}.

The threat landscape has changed dramatically since the introduction of the CARO standard nearly 30 years ago. The quantity of new malware samples that security vendors' labs receive has increased dramatically, to millions per month. Some of these samples are variations of previously known malware, while others take code from older campaigns and re-purpose it for new tasks. Yet still others, are entirely novel types of malware. In this scenario it becomes practically unfeasible to manually and consistently group each malicious file into a well defined hierarchy of families. Even the arguably simpler task of assigning a malware file into an existing family has also become much harder, as malware became more resistant to signature-focused detection, thanks to advanced obfuscation measures such as polymorphism and metamorphism, (repeated) packing and obfuscation, recompilation, and self-updating \cite{rudd2017survey}.

The increasing quantities of malware samples and the resilience to signature methods caused the security community to start using more flexible analytic tools than signatures designed only to identify a single malware variant \cite{gameofthename}, and increasingly rely on dynamic analysis and more generic signatures when possible \cite{adosebyanyothername}. While generic signatures offer an advantage for malware conviction,  the nature of this approach makes it more difficult to organize malware names into families. Moreover, the quantities of malicious files to be analyzed has led to less structured categorizations and greater inconsistencies between vendors. 
In particular, modern security vendor detection names typically fall into one of four categories, containing varying amounts of information about the threat family to which a malware sample belongs.

\emph{Traditional family based}:
Names are associated with unique and distinctive attributes of the malware and its variants. Malware classified under these names usually have a larger amount of original source code or a novel exploit mechanism and often come from the same origins.  This not only gives them a distinctive attribute that can aid in the classification, but also requires researchers to put forth more effort to analyze their inner-workings.  These types of detection names are often of high-quality and have more consistency across vendors.

Today, these detection names are most often seen in parasitic file-infectors and specific botnet campaigns with distinctive attributes, e.g. \emph{Virut}, \emph{Sality}, \emph{Conficker}, etc.  These types of names usually use a suffix to identify specific variants, which often denotes a revision to the malware or change in the configuration data for use in a different campaign. E.g. \emph{Mal/Sality-D}.

\emph{Technique based}:
These types of detection names group together malware that may come from different origins and/or have multiple authors but share a common method or technique. 
For example, many executable \emph{autorun} worms have been written in the past using languages languages like Visual Basic 6 or AutoIt that change explorer's file and folder view settings to hide filename extensions and employ an icon resource similar to that of a popular document format. 
Due to the relatively low complexity of the infection method many amateurs copied this technique resulting in a large amount of similar malware that was not necessarily of the same origin, neither tries to perform the same action in the host machine.

Some anti-malware solutions would generically detect and classify many of these malware samples under the same generic family name, where other vendors may have have defined different more specific criteria for each family classification based on other attributes of the payload.
When a generic family name provided by the AV vendor, they oftentimes replace the detection name suffix with a partial hash of the file data in order to identify a specific sample. The difference in detection methods employed often results in less consistency in detection names across vendors. E.g. \emph{Troj/AutoIt-CHN}.

\emph{Method based}:
This type of detection name simply denotes the detection technology used to detect the malware sample. Some detection names can simply be that of a patented technology, project, or internal code name specific to the AV vendor, indicating the use of heuristics, ML, or real-time detection technologies like cloud look-ups.
In these cases the detection name is not that of a malware family, but that of the method that was utilized to detect the sample. E.g. \emph{Unsafe.AI\_Score\_64\%}.

\emph{Kit based}:
AV vendors will often use more generic family names for detecting malware that has been generated by a known kit.
These kits are often referred to as grey hat tools, as they can be used both offensively by penetration testing teams and by malware authors.
Many of these kits obfuscate their payloads in attempt to circumvent detection by AV software.
Detection names in this category tend to not describe the origins or functionality of the specific malicious payload, but instead identify methods used by the kit or tool to obfuscate or hide their payload. E.g. \emph{Trojan:Win32/Meterpreter.gen!C}. 

In \cite{avclass}, Marcos et al. identified a number of naming inconsistencies across cybersecurity vendors, and proposed AVClass: a malware labeling tool that uses data-mining techniques to distill family names for malware samples by combining detection names from multiple anti-malware solutions. Complementary to AVclass, Perdisci et. al. proposed an automated technique in \cite{perdisci_vamo} which relies on individual detection names from multiple vendors, for evaluating the quality of a given malware clustering. We follow the idea of combining detection names from multiple vendors to better understand the nature of malware samples, but instead of trying to fit each new sample to a naming scheme with mutually exclusive hierarchical categories such as families, we propose an alternate approach to describing the functionality and the relationship between malicious samples by using attribute tags. A set of attribute tags describes a piece of malware through easy to interpret properties, and can be thought of as a soft-family classification, since it describes the sample and relates it with other samples described with the same (or an overlapping) set of tags. The advantage of the tagging approach is that it does not presume a partition on the malware space by genealogy, while providing potentially more actionable information about a malware sample.

The idea of describing malware through a set of descriptive attributes instead of using malware families is not new: The MITRE corporation developed the Malware Attribute Enumeration and Characterization (MAEC) Effort \cite{maec}, a standardized language for attribute-based malware characterization, in version 5.0 at the time of writing. This structured language aims to encode all possible known information about malware based upon an extensive list of attributes such as behaviors, capabilities, artifacts, and relationships between malware samples, among others.
Because the level of detail in the descriptions proposed in MAEC, it becomes prohibitively expensive to characterize a large corpus of malware samples with these attributes. Furthermore, the set of samples that it is possible to characterize with this method is highly biased towards those that can be executed in a controlled sandbox environment.
For the purpose of this work, we decide to work with a reduced, independently defined set of tags. Nevertheless, the techniques we explore throughout this paper apply to broader attribute definitions.

\subsection{Semantic Attribute Tagging}
\label{background:attributes}

Semantic attribute tagging refers to the association of samples with key-words that convey various types of high-level information about their content. These tags can later be used to interpret or summarize the content of the sample, for information retrieval in a large database of samples or for clustering, among others. In the last decade the use of tags has become a popular method for organizing and describing digital information. Content platforms use tags for images, video, audio, news articles, blog posts, and even questions in question-answering forums. 

Automatic content tagging algorithms attempt to annotate data by learning the relationships between the tags and the content. Because any given sample can be related to multiple tags, this task can be, and usually is, framed as a multi-label prediction problem within the field of machine learning. Automatic image \cite{hashtag_prediction, zhang_fast_2016, frome_devise, chen_fast_image_tagging, weston_large_2010,  gong_deep_2013} and audio \cite{choi_automatic_2016, choi_effects_2017, audio_content_based} tagging are among the most popular areas of research in automatic attribute tag prediction today. State of the art text, image and audio tagging algorithms use deep learning techniques which require massive datasets of tagged samples to train on.
These datasets are often generated collaboratively (either directly or indirectly), meaning that multiple sources annotate some part of the dataset independently of each other. As noted by Choi et al. in \cite{choi_effects_2017}, this way of obtaining labeled information is a noisy process which has to be accounted for in the design and evaluation of the learning algorithm. Particularly Choi et al. study the effect of noisy labels when training deep neural networks in the multi-label classification setup, particularly when the noise is skewed towards the negative labels.

Semantic attribute tagging has two important characteristics worth considering: i) it can convey a lot of identifying information about a sample, even if the the sample is novel. Facial attribute tagging \cite{kumar2009attribute, rudd2016moon, wilber2014exemplar, rozsa2016facial, rozsa2017facial, scheirer2012multi}, for example, has repeatedly demonstrated that vectors of attribute predictions (e.g., gender, hair color, ethnicity, etc.) from one or more classifiers can themselves be powerful feature vector representations for face recognition and verification algorithms; ii) semantic tags can be stored, structured, and retrieved in a human interpretable manner \cite{scheirer2012multi}. Both of these characteristics are appealing in a commercial computer security use case where the type of the threat can be roughly identified by a description that makes sense to security researchers and end users.

\subsection{Multi-Label Classification}

We briefly mentioned in Section \ref{background:attributes} that semantic attribute tagging relies on multi-label classification, wherein we aim to predict multiple labels simultaneously. 
There are several ways to do this, the most trivial of which is to learn one classifier per label. This naive approach is not efficient in the sense that one classifier does not benefit from what the other classifiers have learned about a given sample. Furthermore, it can be unfeasible from a deployment perspective, particularly as the number of labels grow.
For correlated labels, a popular approach is to use a single classifier with multiple outputs, one per output label. The total loss for the classifier is obtained by adding together the loss terms across the model's outputs during training and optimizing over a multi-objective loss. Not only does this yield a more compact representation but it also improves classification performance over using independent classifiers \cite{rudd2016moon}. 
We use as our baseline architecture a multi-label deep neural network architecture in Section \ref{sec:tagging_with_dnn} which exploits a shared representation of the input samples, and has multiple binary cross entropy loss functions atop stacks of hidden layers, or heads, with final sigmoid outputs -- one per tag.

An alternative approach to multi-label classification, first introduced in the image tagging and retrieval literature, is to learn a compact shared vector space representation to which map both input samples and labels -- a joint embedding \cite{weston_large_2010, frome_devise, chen_fast_image_tagging, multi_view_embedding2014, hashtag_prediction, zhang_fast_2016} -- where similar content across modalities (images and tags for image tagging) are projected into similar vectors in the same low dimensional space. At query time, a similarity comparison between vectors in this learned latent space is performed, e.g., via inner product, to determine likely labels. A variety of models could be employed to form a joint embedding, but crucially, the embedding is optimized across input modalities/labels. In Section \ref{sec:tagging_with_dnn} we present a joint embedding model that maps malware tags and executable files into the same low dimensional Euclidean space for the malware description problem.

\subsection{Malware Analysis with Neural Networks}

In recent years multiple advances in machine learning for information security (ML-Sec) have taken place. This can be attributed to several factors including an explosion in labeled data available from vendor aggregation services and threat intelligence feeds, and more powerful hardware and software frameworks for fitting highly expressive classifiers, along with a need of the cybersecurity industry to incorporate more flexible methods to improve their detection pipelines. In this work we focus particularly on analysis over Windows Portable Executable (PE) files based on static features, i.e. information that can be extracted from the binary files without having to execute them.

In contrast to our work which focuses on malware description and representation, most modern applications of deep learning have focused on malware detection. Saxe et al. \cite{saxe_berlin} applied deep neural network detection to feature vectors derived from 2-dimensional histogram statistics of PE files along with hashed delimited strings and hashed elements from the file header, including metadata and import tables. Further applications of deep learning exploiting similar feature sets have been used to categorize web content \cite{saxe2018deep}, office documents \cite{rudd2018meade}, and archive formats \cite{rudd2018meade}. Other types of features and classifiers have also been used for the task of PE malware detection. For instance, Raff et al. demonstrated in \cite{minimal_dk} a way to effectively identify malware using solely an embedding of the first 300 bytes from the PE header. In later work, Raff et al. proposed an embedding strategy which takes in the entire PE file \cite{nvidia} for the same problem. In \cite{using_opcodes}, Bugra and Erdogan use a disassembler to retrieve the opcodes of the executable files and then a shallow network based on \textsc{word2vec} \cite{word2vec} to embed them into a continuous vector space. Afterwards, they train a gradient search algorithm based on Gradient Boosting Machines for the malware classification task.

The two approaches that are most related to our work are the works conducted by Huang et al. in \cite{huang2016mtnet} which uses the auxiliary task of predicting family detection names with the goal of improving the performance on their detection model, and by Rudd et al. in \cite{aloha}, work that is contemporaneous to ours, where the authors study the impact of using multiple auxiliary loss terms on a multitude of tasks, one of which is tag prediction, in parallel to the main binary detection task and conclude that using these auxiliary information during training is beneficial for the performance on the main task. Note, however, that the purposes of the auxiliary losses in these works were to improve performance on the main malicious/benign detection task and are not concerned with descriptions of malicious software, similarities between them, or latent representations.

\section{Semantic Malware Attribute Tags}
\label{sec:mal_tags}
We define SMART tags (which we will also refer to as a \emph{malicious} or \emph{malware} tags) as a potentially informative, high-level attributes of malicious or potentially unwanted software. 
These tags are loosely related to malware families, in the sense that they attempt to describe how a piece of malicious software executes and the intent behind it, but they do so in a more general and flexible way. 
One malware campaign (or family) can be associated with more than one tag, and a given tag is associated with multiple families. 
For the purpose of this study, and without loss of generality, we define a set of malicious tags $\mathcal{T}$, with $|\mathcal{T}| = 11$ different tags (or descriptive dimensions) of interest that we can use to describe malicious PE files: \emph{adware}, \emph{crypto-miner}, \emph{downloader}, \emph{dropper}, \emph{file-infector}, \emph{flooder}, \emph{installer}, \emph{packed}, \emph{ransomware}, \emph{spyware}, and \emph{worm}. 
We chose this particular set of tags so that we can generate concise descriptions for most common malware currently found in the wild. The definitions for each of the tags can be found in Appendix \ref{apendix:definitions}.

Since malware tags are defined at a different level of abstraction than malware families, we can bypass the problem of not having a common naming strategy for malicious software, and thus exploit the knowledge contained in multiple genealogies generated from different sources in a quasi-independent manner: detection technologies, methodologies, etc.
It becomes irrelevant if one source identifies a sample as being part of the \emph{Qakbot} family while another calls it \emph{Banking Trojan} so long as we have a way to associate those two correctly with the \emph{spyware} tag \footnote{\emph{Qakbot} in particular also exhibits the behavior of a \emph{worm} and could be therefore also tagged as such.}. Furthermore, this approach allows us to exploit the fact that some sources might have stronger detection rules for certain kinds of malware.

In the remainder of this section we propose a simple labeling strategy used to generate tags at scale for a given set of files that combines information encoded in the detection names of several anti-malware solutions and then translates them into semantic tags.

In later sections we will use this labeled set for both training and evaluation of deep neural networks (DNNs) that annotate previously unseen samples in real time, by solely looking at their binary representation.

\begin{table*}[ht!]
\caption{An example of how our tags are derived from detection names from multiple sources. The first column shows detection names from ten different vendors, where the value \emph{None} indicates that the vendor has not identified the sample as  malicious. In the second column the tokens parsed and normalized from the detection names are listed. The last column shows the tags associated with the tokens in the middle column. This association is represented by using the same color in the tokens and their related tags.
}
\centering{
\begin{tabular}{l|l|l}
\multicolumn{1}{c|}{Detection name} & \multicolumn{1}{c|}{Parsed tokens}         &
\multicolumn{1}{c}{Tags}            \\ \hline \hline
\begin{tabular}[c]{@{}l@{}}Ares!4A26E203524C, Downloader,\\ a variant of Win32/Adware.Adposhel.AM.gen,\\ None, None, None,\\ Gen:Variant.Razy.260309, None,\\ Trojan ( 005153df1 ), Riskware/Adposhel\end{tabular}                                                    & \begin{tabular}[c]{@{}l@{}}ares, {\color{blue} downloader},\\ variant, win32, {\color{red} adware}, {\color{red} adposhel}, gen,\\ \\ gen, variant, razy,\\ trojan, riskware, {\color{red}adposhel}\end{tabular}                                                                         & \begin{tabular}[c]{@{}l@{}}{\color{red} adware}\\ {\color{blue} downloader}\end{tabular} \\ \hline
\begin{tabular}[c]{@{}l@{}}W32.Virlock!inf7, TR/Crypt.ZPACK.Gen,\\ Trojan ( 004d48ee1 ), Virus:Win32/Nabucur.D,\\ W32/VirRnsm-F, Virus.Win32.PolyRansom.k,\\ Win32.Virlock.Gen.8, W32/Virlock.J,\\ Trojan-FNET!CCD9055108A1,\\ a variant of Win32/Virlock.J\end{tabular} & \begin{tabular}[c]{@{}l@{}}w32, {\color{red} virlock}, inf7, tr, {\color{blue} crypt}, {\color{blue}zpack}, gen,\\ trojan, win32, {\color{red}nabucur},\\ w32, vir, {\color{red}rnsm}, {\color{violet}virrnsm}, win32, poly, {\color{red}ransom}, {\color{red}polyransom},\\ win32, {\color{red}virlock}, gen,\\ trojan,\\ variant, win32, {\color{red}virlock}\end{tabular} & \begin{tabular}[c]{@{}l@{}}{\color{red}ransomware}\\ {\color{blue}packed}\\
{\color{violet} file-infector}
\end{tabular} \\\hline
\end{tabular}
}
\label{tab:tag_extraction}
\end{table*}

\subsection{Tag Distillation from Detection Names}
\label{sec:tags_from_detection_names}

High quality tags for malware samples at the scale required to train deep learning models can be prohibitively expensive to create manually. Instead, we rely on semi-automatic strategies that are noisier than manual labeling, but allow us to label millions of files that can then be used to train our classifiers. 
For this purpose, we propose a labeling function that annotates PE files using the previously defined set of tags by combining information contained in detection names from multiple vendors\footnote{Vendor names were anonymized throughout this work to avoid inappropriate comparisons.}. 
In this work we use family names from ten anti-malware solutions that are known to produce high quality detection names, as our starting point.
We note that, using multiple anti-malware vendors is one possible strategy for labeling that leverages expert knowledge about a given sample. Nevertheless, the overall framework we propose can easily accommodate multiple other sources for labeling information, including manual and sandbox-derived (dynamic) analyses.

The labeling process for our experiments consists of three main stages: i) \emph{token extraction}, ii) \emph{token-to-tag mapping}, and iii) \emph{token relationship mining}.
The token extraction phase consists normalizing and parsing the multiple detection names and converting them in sets of sub-strings (bag of tokens representation). In a similar way that the AVClass labeling tool \cite{avclass} does, the token-to-tag mapping stage uses rules that associate relevant tokens with the set of tags of interest. These association rules were created from expert knowledge by a group of malware analysts. Finally, we extend this mapping by mining statistical relationships between tokens to improve tagging 
stability and coverage. Example outputs of each intermediate stage of the tag distillation are represented in Table \ref{tab:tag_extraction}. The full details for the proposed labeling procedure can be found in Appendix \ref{app:labeling}.

The tags obtained with this labeling strategy can be noisy because of the ``crowd-sourcing'' nature of the method used in extracting information (tokens) from multiple quasi-independent sources.
On the other hand, this methodology has the advantage of being cheap to compute and having high coverage over samples, which is critical for developing a high quality ML model.

It is also important to note that this labeling technique generates primarily \emph{positive} relations: meaning that a tag being present identifies a relationship between the sample and the tag, but its absence does not necessarily imply a strong negative relation.

\section{Tags Prediction}
\label{sec:tagging_with_dnn}

\begin{figure}[!t]
    \centering
    \includegraphics[width=0.9\linewidth]{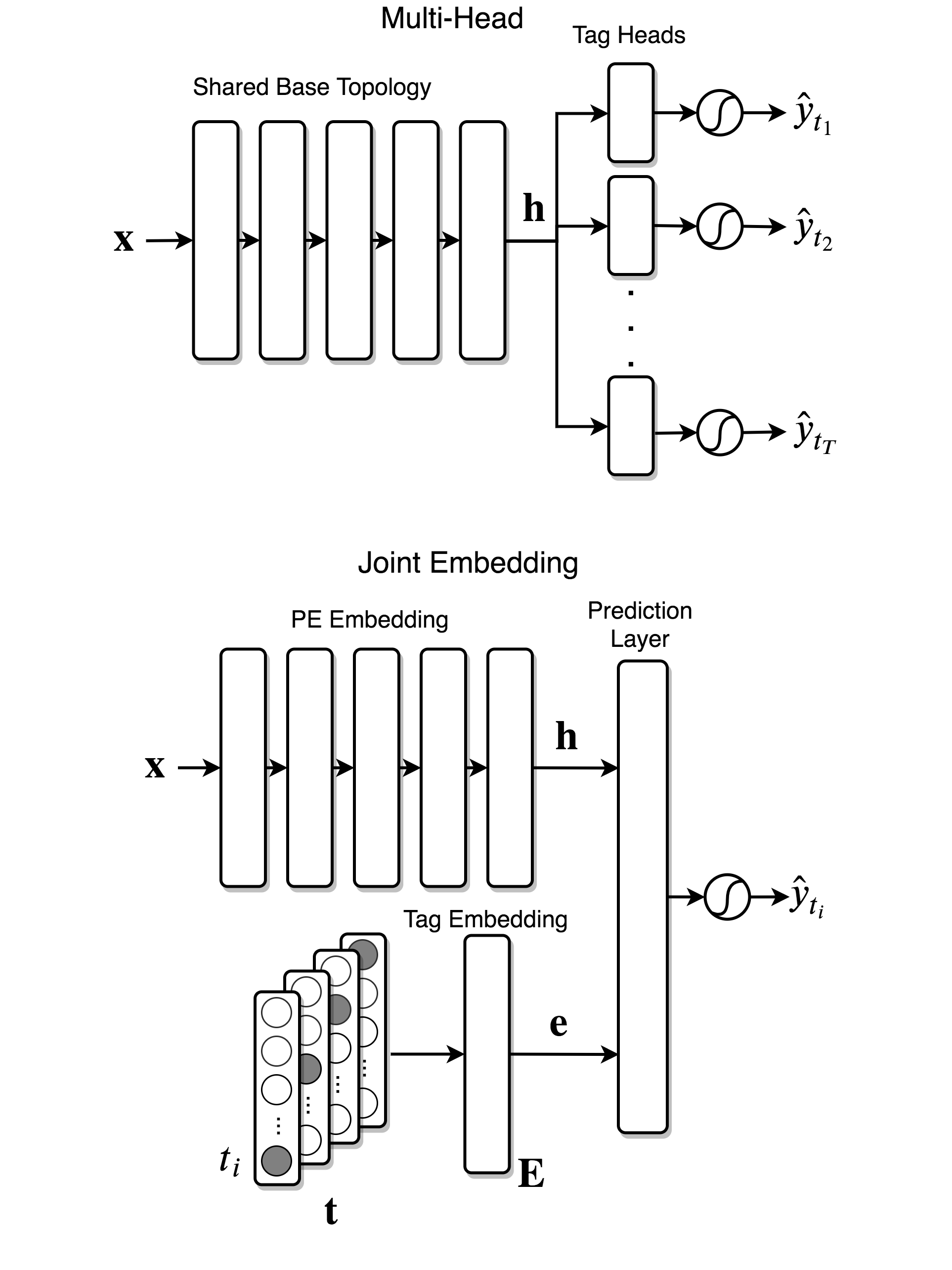}
    \caption{Using samples and corresponding tags we train two neural network architectures to predict malware tags. \emph{Top}: Multi-Head architecture, consisting of a base feed-forward network with one ``head'' for each tag $t_i$ that it is trained to predict. Each of the heads is composed of dense layers followed by ELU nonlinearities, and a final sigmoid activation function. \emph{Bottom}: Joint Embedding model, which represents (embeds) both the binary samples $\mathbf{x}$ and the malicious tags in the same low dimensional space. The prediction layer issues predictions based on the distances between sample embeddings $\mathbf{h}$ and tag embeddings $\mathbf{e}$ in this space.}
    \label{fig:arch_diagrams}
\end{figure}

The goal our work is to flexibly predict important malware qualities from static features, which we more formally define as multi-label classification problem, since zero or more tags from the set of $T$ possible tags  $\mathcal{T} = \{t_1, t_2, \dots, t_{T} \}$ can be present at the same time for a given sample. In order to predict these tags, we propose two different neural network architectures, both represented in Figure \ref{fig:arch_diagrams}, which we will refer to as \emph{Multi-Head} (top) and \emph{Joint Embedding} (bottom).

The \emph{Multi-Head} architecture can be thought as an extension of the network used in \cite{saxe_berlin} to multiple outputs. It consists of a base topology that is common to the prediction of all tags, and one output (or ``head'') per tag.
The base topology can be thought of a feature extraction (or embedding) network that transforms the input features $\mathbf{x}$ into low dimensional hidden vector $\mathbf{h}$, while each head is a binary classifier that predicts the presence or absence of each tag from it.
Both parts of the architecture consist of multiple blocks composed of dropout \cite{srivastava2014dropout}, a dense layer, batch normalization \cite{ioffe2015batch}, and an exponential linear unit (ELU) activation function \cite{clevert2015fast}. The only exceptions are the input layer, which does not use dropout, and the very last layer of each head, which uses a sigmoid activation unit to compute the predicted probability for each label.

The \emph{Joint Embedding} model, as shown at the bottom of Figure \ref{fig:arch_diagrams}, is introduced with three main purposes: i) in an attempt to better model semantic similarities between tags; ii) to explicitly define a low dimensional space that allows us to naturally measure similarities between files; and iii) to have a more flexible architecture that can scale to larger number of tags. 

This model maps both the labels (malware tags) and the binary file features $\mathbf{x}$ to vectors in a joint Euclidean latent space. These embedding functions of files and tags are learnt via stochastic gradient descent in a way such that, for a given similarity function, the transformations of semantically similar labels are close to each other, and the embedding of a binary file should be close to that of its associated labels in the same space.
This architecture consists on a \emph{PE embedding} network, a \emph{tag embedding} matrix $\mathbf{E}$, and a \emph{prediction layer}.

The PE embedding network learns a nonlinear function $\phi_{\theta}(\cdot)$, with parameters $
\theta$ that maps the input binary representation of the PE executable file $\mathbf{x} \in \mathbb{R}^{d}$ into a vector $\mathbf{h} \in \mathbb{R}^{D}$ in low dimensional Euclidean space,
\begin{align}
 \phi_{\theta}(\mathbf{x}): \mathbb{R}^d  \xrightarrow{} \mathbb{R}^D. \nonumber
\end{align}
\noindent
The tag embedding matrix $\mathbf{E} \in \mathbb{R}^{T \times D}$ learns a mapping from a tag $t_n \in \mathcal{T} = \{t_1, \dots, t_{T} \}$, to a distributed representation $\mathbf{e} \in \mathbb{R}^D$ in the joint embedding space,

\begin{align}
    \phi_{E}(t): \{t_1, \dots, t_{T}\} \xrightarrow{} \mathbb{R}^D.   \nonumber
\end{align}
\noindent
In practice, the embedding vector for the tag $t_n$ is simply the $n$-th row of the tag embedding matrix, i.e. $\phi_{E}(t_n) = \mathbf{E}_{n}$. 

Finally, the prediction layer compares both the tag and the sample embeddings ($\mathbf{e}$ and $\mathbf{h}$ respectively) and produces a similarity score that is run through a sigmoid non-linearity to estimate the probability that sample $\mathbf{x}$ is associated with tag $t$ for each $t \in \mathcal{T}$. 
In our final model implementation, the similarity score is the dot product between the embedding vectors. The output of the network $f_n(\mathbf{x}| \theta, E)$ then becomes,

\begin{align}
    \hat{y}_n = f_n(\mathbf{x} | \theta, \mathbf{E}) & = \sigma \left( \langle \phi_E(n) , \phi_{\theta}(\mathbf{x}) \rangle \right) \nonumber \\
                & = \sigma \left(\langle \mathbf{E}_n ,  \mathbf{h} \rangle \right) , 
\end{align}
\noindent
where $\sigma$ is the sigmoid activation function, and $\hat{y}_n$ is the probability estimated by the model of tag $t_n$ being a descriptor for $\mathbf{x}$.

We further constrain the embedding vectors for the tags as suggested in \cite{weston_large_2010}, such that:

\begin{align}
    ||\mathbf{E}_n||_2 \leq C, \quad n = 1, \dots, T,
\end{align}
\noindent
which acts as a regularizer for the model. We observed in practice that this normalization indeed leads to better results on the validation set. Unless stated differently we fixed the value of $C$ to 1.

We also experimented with constraining the norm of the PE embeddings to 1, and analogously using cosine similarity instead of a dot product as a similarity score between tags' and files' embeddings.
In both cases we observed deteriorated performance on the validation set. 
This drop in performance was more noticeable for those samples with multiple tags (more than 4), suggesting that the network is using the magnitude of the PE embedding vector to achieve high similarity scores for multiple tags concurrently. 
As part of our experimentation we also tried to learn the similarity score by concatenating together the PE and tag embeddings and running the resulting vector through feed forward layers with nonlinearities. However, we found that that the simpler approach of using dot product was both more effective on the tag prediction task and also lead to easier to interpret predictions.

Our goal, for a given PE file, is to learn a distributed, low dimensional representation of it, that is both ``close'' to the embedding of the tags that describe it and to other PE files with similar characteristics. The parameters of both embedding functions $\phi_{\theta}(\cdot)$ and $\phi_E(\cdot)$ are jointly optimized to minimize the binary cross-entropy loss for the prediction of each tag via backpropagation and stochastic gradient descent. The loss function to minimize for a mini-batch of $M$ samples becomes:

\begin{align}
    \mathcal{L} &= - \frac{1}{M}
        \sum_{i=1}^M
        \sum_{n=1}^{T}
        f_n(\mathbf{x}^{(i)}|\theta, \mathbf{E}) 
          \log (t_n^{(i)})  \nonumber \\ 
        & \qquad \qquad \qquad + (1- f_n(\mathbf{x}^{(i)}|\theta, \mathbf{E}))   \log(1 - t_n^{(i)})  \nonumber \\
        &= - \frac{1}{M}
        \sum_{i=1}^M
        \sum_{n=1}^{T}
        \hat{y}_n^{(i)} \log (t_n^{(i)}) + 
        (1-\hat{y}_n^{(i)}) \log(1 - t_n^{(i)})
\end{align}
\label{eq:loss}

\noindent
where $t_n^{(i)}=1$ if sample $i$ is labeled with tag $t_n$ or zero otherwise, and $\hat{y}_n^{(i)}$ is the probability predicted by the network of that tag being associated with the $i$-th sample.

In practice, to get a vector of tag similarities for a given sample $\mathbf{x}$ with PE embedding vector $\mathbf{h}$ we multiply the matrix of tag embeddings $\mathbf{E} \in \mathbb{R}^{T \times D}$ by $\mathbf{h} \in \mathbb{R}^{D}$ and scale the output to obtain a prediction vector $\hat{\mathbf{y}} = \sigma ( \mathbf{E} \cdot \mathbf{h}) \in \mathbb{R}^T$, where $\sigma$ is the element-wise sigmoid function for transforming the similarity values into a valid probability.
Each element in $\hat{\mathbf{y}}$ is then the predicted probability for each tag.

We note that, because this model maps every tag into the same low dimensional space, it can explicitly account for similarities on the labels. That is, if two tags are likely to occur together, they will be mapped close to each other in embedding space. With this architecture, it is possible to train the model without the need of computing the loss for all negative labels using techniques such as negative sampling, where only positive labels and a randomly selected subset of negative labels are used in the inner summation of Equation \ref{eq:loss}. It would be possible to use a ranking loss as proposed in \cite{weston_large_2010}. We defer the study of these optimizations for future work.

Finally, if we compare both models in Figure \ref{fig:arch_diagrams}, they are mathematically similar, with the main differences being the explicit modeling of the tag embedding step (and its regularization), the subsequent explicit modeling of the joint embedding space that allows us to naturally perform label-to-label and sample-to-sample similarity searches, and the ability to use various distance functions in the prediction layer.

\subsection{Evaluation of Tagging Algorithms}
\label{sec:eval_tags}
There are different ways to evaluate the performance of tagging algorithms. 
Particularly, the evaluation can be done in a \emph{per-tag} or a \emph{per-sample} dimension. The former seeks to quantify how well our tagging algorithm performs on identifying each tag, while the latter focuses on the quality of the predictions for each sample instead.

In the \emph{per-tag} case, one suitable way to evaluate the performance of the model is to measure the area under the receiver operating characteristic curve (AUC-ROC, or simply AUC) for each of the tags being predicted. A ROC curve is created by plotting the true positive rate (TPR) against the false positive rate (FPR). Also, since the target value for the $n$-th tag of a given sample is a binary True/False value ($t_n \in \{0, 1\}$), binary classification evaluation metrics such as `Accuracy', `Precision', `Recall', and `F-score' also apply. To compute these metrics, the output probability prediction needs to be binarized. For the binarization of our predictions, we choose a threshold independently for each tag such that the FPR in the validation set is 0.01 and use the resulting 0/1 predictions. The fact that our labeling methodology introduces label noise -- mostly associated with negative labels, as pointed out in Section \ref{sec:mal_tags} -- makes \emph{recall} the most adequate of these last four metrics to evaluate our tagging algorithms, since it ignores incorrect negative labels.

The \emph{per-sample} evaluation dimension seeks to evaluate the performance of a tagging algorithm for a given sample, across all tags. Let $T^{(i)}$ be the set of tags associated with sample $i$ and $\hat{T}^{(i)}$ the set of tags predicted for the same sample after binarizing the predictions. We can use the Jaccard similarity (or index) $J(T^{(i)}, \hat{T}^{(i)})$ as a figure of how similar both sets are. 
Furthermore, let $\mathbf{y} \in \{0, 1\}^{T}$ be the binary target vector for a PE file, where $\mathbf{y}_n$ indicates whether the $n$-th tag applies to the file and $\hat{\mathbf{y}}$ be the binarized prediction vector from a given tagging model. We define the per-sample accuracy as the percentage of samples for which the target vector is equal to the prediction vector, i.e., all  tags correctly predicted, or, in other words, the Hamming distance between the two vectors is zero. For an evaluation dataset with $M$ samples we can use,

\begin{align}
    \text{Mean Jaccard similarity} &= \frac{1}{M} \sum_{i=1}^M
        J(T^{(i)}, \hat{T}^{(i)}) \nonumber \\
        &=  \frac{1}{M} \sum_{i=1}^M \frac{T^{(i)} \cap \hat{T}^{(i)}}{T^{(i)} \cup \hat{T}^{(i)}} \label{eq:jaccard}\\
    \text{Mean per-sample accuracy} &= \frac{1}{M} \sum_{i=1}^M
    \mathbb{I}(\mathbf{y}^{(i)} = \hat{\mathbf{y}}^{(i)}) \label{eq:accuracy}
\end{align}
\noindent
as our per-sample performance metrics for the tagging problem, where $\mathbb{I}$ is the indicator function which is $1$ if the condition in the argument is true, and zero otherwise.

\section{Experiments}
\label{sec:experiments}
We trained the two proposed model architectures on the task of malware tagging from static analysis of binary files. After training different model sizes with different hyper-parameters in a smaller scale dataset (details below) and evaluating in a validation set, we trained our final network architectures (those model configurations that performed the best) using a large scale dataset to produce our final evaluation results. Finally, we evaluate the quality of the embedding spaces learned by both models as a way of determining similarities between malware files, and compare them with the original feature representation.

In this section we provide the experimental details of this process: particularly a description and analysis of the data used for training and validation along with a definition of the model topology and training methodology.

\subsection{Data Description}
\label{sec:data_description}

For training purposes we collected two different sets, a medium size training set containing 10 million unique binary Windows Portable Executable (PE) files ($D_\text{train-M}$), and a large training set containing 76,204,855 unique PE files ($D_\text{train-XL}$), such that $D_{\text{train-M}} \subset D_\text{train-XL}$.
The smaller training set was used for experimenting with different architectures and validation purposes, while the larger one was used to train our final models, whose architectures performed the best in the validation set when trained in the smaller dataset.
These sets were obtained by random sampling of files first observed in our intelligence feed in the period between 06/25/2018 and 05/26/2019, but not seen after that. Similarly, we created a validation set, $D_{\text{val}}$ composed of 3,159,377 PE files, randomly sampled from the samples seen during the period of one month after the the training set (05/26/2019 to 06/26/2019). We used this set to compare the performance of different candidate models and architectures.

For the final evaluation we sampled 3,456,288 malware samples first seen in the threat intelligence feed between 07/01/2019 and 07/29/2019. This test set, $D_{\text{test}}$ is composed solely of malware samples to better understand the performance of our models when implemented in a real-world scenario of malware description.

For all $D_\text{train-M}$ and $D_{\text{train-XL}}$, $D_{\text{val}}$ and $D_{\text{test}}$ we derived the semantic tags following the procedure described in Section \ref{sec:tags_from_detection_names} and detailed in Appendix \ref{app:labeling}, using detection names from ten anti-malware solutions that we consider provide high-quality names. 
The set of tokens and mappings used was based only on detection names from samples in $D_\text{train}$, in order to avoid polluting our time split evaluation.
We further derived a malicious/benign label for the samples in those sets using a voting scheme similar to \cite{saxe_berlin}, but extended to assign more importance to trusted vendors, and complemented with internal proprietary reputation scores, white and black lists.
The large training set is composed of 79\% malware samples and 21\% benign samples. This ratio is similar for the small training set as well as the validation set.

For all the binary files in the three datasets we then extracted $1024$-element feature vectors using the same feature representation as proposed in \cite{saxe_berlin}, which uses windowed byte statistics, 2-dimensional histograms of delimited string hash vs. length, and histograms of hashes of PE-format specific metadata such as imports from the import address table.

Table \ref{tab:tag_coverage} summarizes the coverage for each of the tags across our train dataset $D_{\text{train}}$.
Most of our tags are almost exclusively associated with malicious samples, except for \emph{installer} and \emph{packed} which are associated with both benign and malicious files. Moreover, we see that 90\% of the malicious samples have at least one tag describing them, indicating that the labeling approach has effectively a high coverage over the set of samples of interest. The mean number of tokens observed for each time that a tag appears is 5.57, which represents the degree of robustness of our labeling strategy against vendor mis-classifications or missing scans. Statistical synonym and parent-child relationships used to produce the tags were computed from the samples in the train dataset. Using both synonym and parent-child relationships derived from the empirical conditional probabilities of tokens improves not only the mean token redundancy but also the tag coverage for malicious samples for almost all our tags, leaving unaffected the tagging for benign samples. The coverage statistics for the validation and test sets are similar to the ones presented in the table and not shown here for space considerations.

\begin{table}[t!]
\caption{Tag coverage, i.e. percentage of samples annotated with a given tag, for the train dataset $D_{\text{train-XL}}$ (as described in Section \ref{sec:data_description}) for benign and malicious files. The last row considers a sample as labeled if any one of the tags is present.}

\centering
\begin{tabular}{l|r|r}
    \multicolumn{1}{c|}{Tag}           
    &  \multicolumn{1}{c|}{\begin{tabular}[c]{@{}c@{}}Benign\\ Samples\end{tabular}}
	&  \multicolumn{1}{c}{\begin{tabular}[c]{@{}c@{}}Malware\\ samples\end{tabular}}	\\ \hline \hline
adware        & 851 ($<$ 0.01\%)      & 12,915,945 (21.46 \%)   \\ \hline
crypto-miner  & 77 ($<$ 0.01\%)       & 1,739,261 (2.90 \%)		\\ \hline
downloader    & 371 ($<$ 0.01\%)      & 13,612,093 (22.61 \%)   \\ \hline
dropper       & 117 ($<$ 0.01\%)      & 18,321,174 (30.44 \%)   \\ \hline
file-infector & 51 ($<$ 0.01\%)       & 15,417,637 (25.61 \%)   \\ \hline
flooder       & 0 (0\%)	              & 602,751  (1.00 \%)		    \\ \hline
installer     & 562,607 (3.51 \%)     & 5,959,342 (9.90 \%)     \\ \hline
packed        & 576618 (3.60 \%)      & 22,517,435 (37.41 \%)   \\ \hline
ransomware    & 10 ($<$ 0.01 \%)      & 6,192,555  (6.25 \%)    \\ \hline
spyware       & 345 ($<$ 0.01\%)      & 22,834,226 (37.93 \%)   \\ \hline
worm          & 56 ($<$ 0.01 \%)      & 15,721,510 (26.11 \%)   \\ \hline \hline
ANY           & 1,067,876 (6.67 \%)   & 54,456,845  (90.47) \%   
\end{tabular}
\label{tab:tag_coverage}
\end{table}

\begin{figure}[ht]
    \centering
    \includegraphics[width=0.85\linewidth]{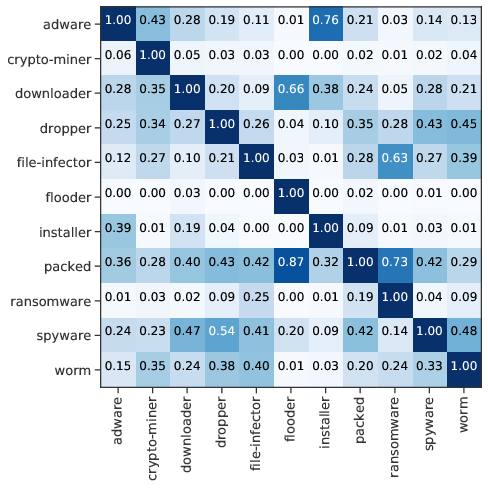}
    \caption{Estimated tag conditional probabilities for our training set $D_{\text{train-XL}}$. The value of the element in the $i$-th row and $j$-th column represents the empirical conditional probability of tag $i$ given tag $j$, $\tilde{p}(t_i|t_j)$ for our labeling strategy.}
    \label{fig:tags_cond_prob}
\end{figure}

We further analyze the distribution and pairwise relationships of the tags in our training dataset. In Figure \ref{fig:tags_cond_prob} we plot the empirical conditional probability of the tags in the train set computed in a similar fashion as in Equation \ref{eq:token_cond_prob} by replacing token counts with tag counts. The value in row $i$ and column $j$ represents the empirical conditional probability of the tag $t_i$ given that tag $t_j$ is present, $\tilde{p}(t_i|t_j)$.
This representation is useful to identify possible issues with the labeling mechanism as well as understanding the distribution of our tags. Furthermore, we can compare this matrix generated on the test set with the one derived from the predictions of the model (instead of the labels), to have a better understanding of the errors that the model is making.

\subsection{Training Details}
\label{sec:training_details}
We trained the two model architectures introduced in Section \ref{sec:tagging_with_dnn} on the training dataset $D_{\text{train-M}}$ with different number of layers and number of nodes per layer, each for 200 epochs using an Adam optimization procedure \cite{adam} on mini-batches of 4096 samples and at a learning rate of $5 \cdot 10^{-4}$.
The combination of layer sizes and number of layers that performed best when trained in the smaller training set where then trained in the large training set $D_{\text{train-XL}}$.

The shared base topology of the final Multi-Head architecture consists of an input feed-forward layer of output size 2048, followed by a batch normalization layer, an ELU nonlinearity and three blocks, each composed by dropout, a linear layer, batch normalization and ELU of output sizes 512, 128, and 32 respectively. Each output head is simply a linear layer (the same for each head) composed of the same type of basic blocks as are in the main base architecture, but with output size 11 (number of tags being predicted) and a sigmoid non-linearity instead of the ELU. Binary cross-entropy loss is computed at the output of each head and then added together to form the final loss.

The Joint Embedding architecture uses the same base topology as the Multi-Head model for the embedding of the PE files into a 32 dimensional joint latent space. An embedding matrix ($\mathbf{E}$) of learnable parameters with size $T \times 32$ is used for the embedding of the tags. We used dot product to compute the similarity between the PE file embedding and the tag embedding followed by a sigmoid non-linearity to produce an output probability score. As before, the sum of the per-tag binary cross-entropy losses is used as the mini-batch loss during model training.

We found experimentally, for both network architectures, that having a wide input layer, i.e. increasing the number of dimensions in the first layer from an input representation of 1024 to 2048 dimensions was crucial to obtaining good performance in the tag prediction task.

\section{Tagging Results}
\label{sec:results}
As mentioned in Section \ref{sec:eval_tags} there are two main dimensions of interest when analyzing the performance of malware tagging algorithm: a \emph{per-tag} dimension, which evaluates how well each tag is predicted and a \emph{per-sample} dimension, which focuses on how many samples are correctly predicted and how accurate those predictions are. In the following we analyze the performance of our models across these dimensions after training in the large scale dataset $D_{\text{train-XL}}$ and evaluating in the test dataset $D_{\text{test}}$.

The evaluation results presented in this section take into consideration only those samples identified as malware by our labeling scheme. This is because the goal of our current tagging algorithm is to describe only malicious or potentially unwanted behaviors. At deployment time, we assume that the tagging models analyze samples already convicted by a complementary mechanism, and so we only evaluate on actual malware to resemble this deployment scenario. Only evaluating in malicious samples also allows us to compare results across different test sets that might not have the same malware/benign ratio of samples.
In Appendix \ref{sec:eval_with_benign} we complement this results by evaluating the performance on a version of the test set $D_{\text{test}}$ that also contains benign samples, and show that the models' performance does not degrade in the presence of them, meaning that the trained models do not compulsory assign malicious tags to benign samples.

\subsection{Per-Tag Results}

\begin{figure}[ht]
    \centering
    \includegraphics[width=0.75\linewidth]{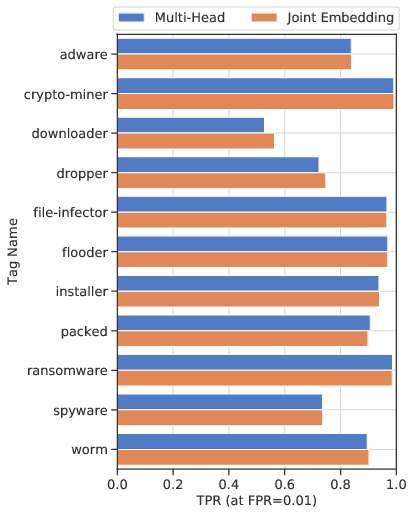}
    \caption{Per-tag true positive rate (TPR) at a false positive rate (FPR) of $10^{-2}$ for the two proposed models evaluated on $D_{\text{test}}$. The Joint Embedding architecture outperforms the Multi-Head architecture by a small margin for all the tags except for \emph{packed} and \emph{flooder}.}
    \label{fig:recall_barcharts}
\end{figure}

After training the two proposed architectures we proceed to evaluate their performance on the test set $D_{\text{test}}$.
In Figure \ref{fig:recall_barcharts} we compare the per-tag true positive rate (TPR or recall) of both the Multi-Head and Joint Embedding architectures at a per-tag false positive rate (FPR) of 1\%. 
The Joint Embedding architecture outperforms the baseline Multi-Head model by a small margin for all the tags except for \emph{packed} and \emph{flooder} for which the latter performs slightly better. The largest improvements over the baseline appear for the \emph{downloader} and \emph{dropper} tags, for which the Joint Embedding model outperforms the baseline by 3.6\% and 2.4\% respectively.
We have observed this trend of obtaining better results with the Joint Embedding architecture consistently for other experiments -- with different datasets, layer sizes, and activation functions -- that we carried out during the development of this study.

Table \ref{tab:pertag_results} provides a more thorough comparison of these two methods. 
Not only does the Joint Embedding model outperforms the baseline in terms of mean recall, but it also does so in terms of AUC for every tag except \emph{adware}. For computing both recall and F-score we binarized the output using a threshold such that the FPR in the test set is 1\% for each tag. For these two binary classification metrics, the Joint Embedding model exhibits better performance than the Multi-Head model on average: the Multi-Head architecture achieves a mean recall of 0.862 and mean F-score of 0.837 while the proposed Joint Embedding model achieves an average recall of 0.867 and a F-score of 0.841.

\begin{table*}[t!]
\caption{Per-tag evaluation results for the two proposed architectures on the malware samples of the test dataset ($D_{\text{test}}$). Both recall and F-score are computed by binarizing each classifier’s outputs at a false positive rate of 0.01 on the test set for each tag. The weighted mean weights the contribution of each tag by its support. The best result between the two proposed architectures for each row is highlighted in bold}
\centering
\begin{tabular}{l|ccc|ccc}
\multicolumn{1}{c|}{} & \multicolumn{3}{|c|}{Multi-Head}  & \multicolumn{3}{c}{Joint Embedding} \\
Tag name        & AUC               & \begin{tabular}[c]{@{}c@{}}Recall \\ @FPR=$10^{-2}$ \end{tabular}   & \begin{tabular}[c]{@{}c@{}} F-score \\ @FPR=$10^{-2}$ \end{tabular}
                & AUC               & \begin{tabular}[c]{@{}c@{}}Recall \\ @FPR=$10^{-2}$ \end{tabular}   & \begin{tabular}[c]{@{}c@{}} F-score \\ @FPR=$10^{-2}$ \end{tabular}         \\ \hline \hline
adware          &  \textbf{0.983}        & \textbf{0.839}          & \textbf{0.861} 
                &  0.982                 & \textbf{0.839}          & \textbf{0.861}         \\
crypto-miner    &  \textbf{0.999}        & 0.991                   & \textbf{0.865} 
                &  \textbf{0.999}        & \textbf{0.992}          & \textbf{0.865}         \\
downloader      &  0.982                 & 0.527                   & 0.676 
                &  \textbf{0.984}        & \textbf{0.563}          & \textbf{0.706}         \\
dropper         &  0.975                 & 0.723                   & 0.829  
                &  \textbf{0.976}        & \textbf{0.747}          & \textbf{0.845}         \\
file-infector   &  0.996                 & \textbf{0.966}          & \textbf{0.968} 
                &  \textbf{0.997}        & \textbf{0.966}          & \textbf{0.968}         \\
flooder         &  0.985                 & \textbf{0.970}          & \textbf{0.546} 
                &  \textbf{0.988}        & 0.969                   & \textbf{0.546}         \\
installer       &  \textbf{0.994}        & 0.938                   & 0.809 
                &  \textbf{0.994}        & \textbf{0.940}          & \textbf{0.810}         \\
packed          &  \textbf{0.993}        & \textbf{0.907}          & \textbf{0.943} 
                &  \textbf{0.993}        & 0.899                   & 0.939                  \\
ransomware      &  \textbf{0.998}        & \textbf{0.986}          & \textbf{0.943} 
                &  \textbf{0.998}        & \textbf{0.986}          & \textbf{0.943}         \\
spyware         &  0.977                 & \textbf{0.736}          & 0.841 
                &  \textbf{0.979}        & \textbf{0.736}          & \textbf{0.842}         \\
worm            &  0.989                 & 0.896                   & 0.927 
                &  \textbf{0.991}        & \textbf{0.902}          & \textbf{0.931}         \\ \hline
mean            &  0.988                 & 0.862                   & 0.837 
                &  \textbf{0.989}        & \textbf{0.867}          & \textbf{0.841}        \\
weighted mean   &  0.985                 & 0.809                   & 0.868 
                &  \textbf{0.986}        & \textbf{0.816}          & \textbf{0.874}
\end{tabular}
\label{tab:pertag_results}
\end{table*}

\subsection{Per-Sample Results}

Another way of analyzing our results is to measure the percentage of samples for which our models accurately predicted all tags. We are also interested in knowing how many tags on average (out of the 11 possible tags) each model correctly predicts per sample.
For this we measure both the Jaccard similarity and the per-sample accuracy of our predictions according to equations \ref{eq:jaccard} and \ref{eq:accuracy} respectively. Under both metrics the Joint Embedding model outperforms the Multi-Head approach. For the Joint Embedding architecture, the average number of samples for which we predict \textit{all} the tags correctly is 66.8\% while if we choose a sample at random, the model correctly predicts the presence (and absence) of each tag for almost 96\% of the tags (10.53 over 11 possible tags correctly predicted) on average. It is important to note that, because of the relatively low number of tags per sample -- a mean of 2.26 in $D_{\text{test}}$ -- the mean Jaccard similarity for a tagging algorithm that never predicts any tag would be 79.4\% in this test set. Even though this baseline is already high, both our tagging models outperform it by a large margin, which signals that the models are effectively learning to identify relationships between tags and binary feature vectors.

\begin{table}[ht]
\caption{Evaluation results for our two proposed architectures on the malicious samples of the test set, $D_{\text{test}}$. Jaccard similarity and accuracy are both computed according to  equations \ref{eq:jaccard} and \ref{eq:accuracy} respectively. In both cases the performance of the Joint Embedding model outperforms the baseline by a noticeable margin.}
\centering{
\begin{tabular}{l|c|c}
    		& Multi-Head (XL)		& Joint Embedding (XL)  \\ \hline\hline
\begin{tabular}[c]{@{}l@{}}Sample-wise\\ Accuracy\end{tabular} 
            & 0.657		   				& \textbf{0.668}		\\ \hline
\begin{tabular}[c]{@{}l@{}}Jaccard\\ Similarity\end{tabular}
            & 0.956   				& \textbf{0.957}
\end{tabular}
}
\label{tab:persample_results}
\end{table}

\section{Latent Space: Analysis and Applications}
\label{sec:discussion}
Our results from Section \ref{sec:results} show that the Joint Embedding approach is more suitable for malware tagging than the Multi-Head model architecture. Because the number of parameters in both networks is comparable, it means that the Joint Embedding network is learning a more informative internal representation of the binary files, guided by the ability to model, and therefore exploit tag relationships (labels structure) in the latent space. 

This improved internal representation has, in turn, two main advantages compared with the original feature representation: i) it is highly compressed (32 dimensions versus 1024) which allows for indexing malware databases in a more efficient way, and perform similarity computations and nearest neighbor searches faster; and ii) this representation also provides a more informative space where to measure files' similarities in terms of malicious capabilities that is aligned with the original underlying family structure of malware.

In Section \ref{sec:tag-mal_embeddings}, we verify that the Joint Embedding model has learned a proper representation by examining its latent space and validating that PE file embeddings tend to cluster around their corresponding tag embeddings. We then explore the practical applications of having such a representation space within the context of information retrieval.
In Section \ref{sec:embedding_comparison} we compare the learnt representation space by the Joint Embedding model, with the intermediate 32 dimensional representation of the Multi-Head model (output of the Shared Base Topology, see Figure \ref{fig:arch_diagrams} top figure), and the input features, in terms of their ability to represent malware characteristics. We found that the more informed representation learned by the Joint Embedding model can be used to identify novel malware campaigns for which we may have a very small set of samples.

\subsection{Malware-Tag Joint Embedding Space}
\label{sec:tag-mal_embeddings}

\begin{figure}[ht]
    \centering
    \includegraphics[width=0.9\linewidth]{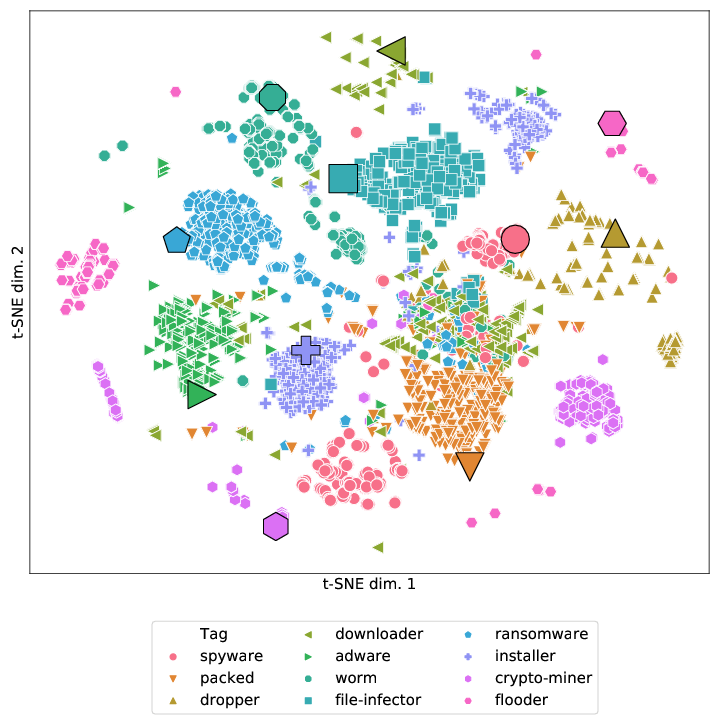}
    \caption{T-SNE visualization of sample and tag (label) embeddings for 8,250 samples labeled with a single tag from $D_{\text{test}}$ (750 randomly selected samples per tag). Large markers represent tag (label) embeddings while small markers represent PE samples embeddings. Samples with the same labels tend to cluster together, and at the same time around their corresponding tag embeddings.}
    \label{fig:tag_sample_embedding}
\end{figure}

As a way to validate and understand the latent space learned by our Joint Embedding model we used t-SNE \cite{tsne} to reduce the dimensionality of the 32-dimensional latent space to a 2-dimensional representation as shown in Figure \ref{fig:tag_sample_embedding}. In this visualization, large markers correspond to the embeddings of the tags themselves while small markers are the transformed embedding vectors for a subset of samples in the test set. Particularly, we randomly selected 750 samples annotated with a single tag for every tag, for a total of 8,250 test samples.

As one can see, the embeddings of the PE files labeled with the same tag tend to cluster together. Furthermore, the embeddings of the tag labels lie close to the clusters of samples they describe. This suggests that the Joint Embedding model has effectively learned to group file representations close to their corresponding tag representations, as intended.

This structure in embedding space is a powerful representation for information retrieval (IR) applications. The task of malware tagging itself can be thought of as a particular case of IR, in which we we retrieve descriptive information for a query point (binary file) based on a distance function in latent space. 
Similarly, it is also be possible to perform file similarity searches: using one malware sample to retrieve other samples with similar malicious characteristics by simply retrieving files for which the embedding representation is close to the representation of the query sample. We explore how this can be used in the context of identifying samples from novel campaigns in  Section \ref{sec:embedding_comparison}. 
Furthermore, it can also be used for obtaining examples of malicious programs that fit a given tag description, where with given a combination of descriptive tags we can define a subspace within the embedding space containing those samples that closely match that particular tag combination of interest.

\subsection{File-File Similarity Index}
\label{sec:embedding_comparison}

The proposed representation of PE files in embedding space, along with the distance function defined in this space, allows us to measure not only similarities between files' and tags' embeddings, but between files' representations as well.

Because of how our models were trained, two pieces of malware that are close to each other in latent space should exhibit similar malicious capabilities. This means that it is possible to use the latent representation of the files to measure how similar two pieces of malware are in terms of their capabilities. Moreover, the size of the file embedding representation (32 dimensions) is 32 times smaller than the original file feature representation (1024 dimensions). This makes the task of storing, indexing and querying large databases of malware more efficient.

In this section we evaluate the degree to which the embedding representation is more informative in terms of malware attributes when compared with the original 1024-dimensional feature representation.
If the learned embedding function is effectively modeling malicious characteristics of the executable files, then a similarity analysis of files in this space should be more informative in terms of malware attributes' similarity than the original feature space. In other words, because we are indirectly training our models with malware families' information generated by a group of (quasi-)independent ``experts'' (anti-malware vendors), via tagging decomposition, then we expect to be able to differentiate malware campaigns from each other, even if they are described by the same set of tags.

We set to evaluate this, by first identifying well known malware families in the test set. To do so, we rely on a proprietary behavioral sandbox environment where we run a subset of the samples from the test set and look for specific patterns, known via reverse engineering, to be associated with notable malware campaigns. This patterns come from information in memory dumps, network traffic packet captures, file read and write operations, as well as many other activities that would not necessarily be observable with a static scan alone (and thus information not available to our models), since in its binary state, this data could be encrypted, or possibly not even present (as it may be downloadded at runtime).

With this method, we identified the top $f = 13$ most prominent malware campaigns in the test set that can be convicted with high confidence. Table \ref{tab:mal_fams} in the Appendix lists the family names as well as their associated tags and counts in the test set. We note that there are a number of families with the same tag description, for instance \emph{remcos} and \emph{emotet} families are both described with the \emph{sypware} tag.

Next, we created the joint embedding representation for these files using our already trained Joint Embedding network. For each file, we also created a continuous vector representation from the trained Multi-Head model by simply computing the output of the last layer (after the ELU activation) of the shared base topology, represented by $\mathbf{h}$ in the top diagram of Figure \ref{fig:arch_diagrams}.

Finally, we set up the task of $f$-way family classification via nearest neighbor search. It consists in randomly sampling $k$ files per family as anchor samples, and $q$ files per family as query samples. Each of the ($f \cdot q$) query samples is then predicted to belong to the same class as its closest anchor sample in feature space. A more descriptive representational space should separate malware families, i.e. pull those samples that belong to the same family close to each other while pushing away samples from other families, thus performing better on the nearest neighbor classification task.

We consider the neural network embeddings and the original 1024 features as feature spaces for this task. We randomly choose 23 query samples per family and vary the number of anchor samples $k$ from 1 to 10. We repeat the sampling process (for both anchors and query samples), classification and evaluation 15 times per value of $k$ to obtain uncertainty estimates for our accuracy results.

\begin{figure}[ht]
    \centering
    \includegraphics[width=0.85\linewidth]{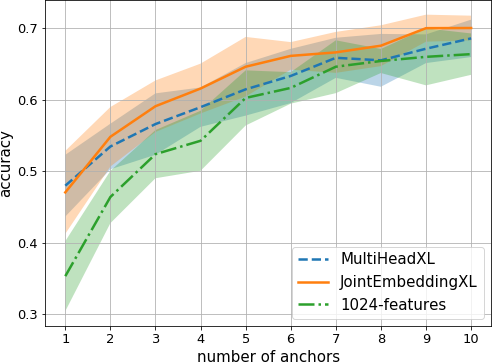}
    \caption{Accuracy results of the $f$-way malware campaign identification via nearest neighbor on input feature space (1024-features) and embedding spaces: both with Joint Embedding and Multi-Head networks trained with the task of tag prediction. By measuring file similarities in embedding spaces we can better identify new samples of a given malware campaign with only a small set of exemplary (anchor) samples (1 to 10 in the figure).}
    \label{fig:kneigh_results}
\end{figure}

Figure \ref{fig:kneigh_results} shows the $f$-way accuracy results for the 3 feature representations: derived from the Joint Embedding model (full orange line), derived from the Multi-Head model (dashed blue line) and computed directly from the binary files (same features that are the input to the neural network models during training -- dotted green line). The baseline accuracy for a model that randomly picks a class is $7.7\%$, independently of the number of anchors. We see that the representation learnt by both the Joint Embedding and Multi-Head models is more effective in identifying family membership than the original set of features.

When the number of anchors is smaller, the mean accuracy for the neural embedding approaches outperforms the input features by a large margin. By only having access to a single malware sample from a given family (the case of one anchor) our approach accurately identifies, on average, 47\% of the remaining samples from that family compared to 36\% identified with the raw feature representation (more than 30\% accuracy improvement).
As the number of anchors grows, the task becomes simpler due to the fact that there is a higher likelihood of having an anchor that is a near duplicate of the query sample, and the improvements over the baseline smaller, although still noticeable.
Particularly, the similarity metric from the Joint Embedding model performs slightly better than the one derived from the Multi-Head network representations.

These results imply that, by learning a representation of malware files in a low dimensional space with a neural network informed by tagging labels, we not only gain the ability to perform tagging on new samples directly from the binary files, but can use it to cheaply compute files' similarities in a low dimensional space, based on their malicious characteristics.

This file similarity metric can then be effectively used in one or few-shot learning scenarios. This is incredibly useful when dealing with novel malware campaigns, as it provides an efficient and fast way of identifying files belonging to a new, recently introduced family for which we only have a limited number of samples (or even one).

\section{Conclusion}
\label{sec:conclusion}

In this paper we first introduced a novel method for malware description via automatic tagging with deep neural networks that learns to characterize malicious software by combining knowledge from multiple quasi-independent sources. With the proposed Joint Embedding model it is possible to accurately predict user interpretable attribute and behavioral descriptions of malicious files from static features, correctly predicting an average of more than 10.52 out of 11 tag descriptors per sample at a per-tag false positive rate of 0.01.

We then analyzed the embedding space learned by our neural networks trained on the task of malware description and showed that the files' representation in this space is more informative in terms of malicious capabilities, and also more compact, than the input feature space used to train the networks initially.

Finally, we introduced a file-to-file similarity index that is highly compressed (32 times with respect to the original feature representation), that allows us to perform more effective and better informed similarity searches over large malware databases along interpretable dimensions of malicious capabilities. We showed how this new metric can be used effectively in the context of few and one shot learning in which we aim to identify specimens of a new malware campaign. When only one example of a given malware family is available, our similarity index improves family classification accuracy by more than 30\% over the baseline features.
We also proposed a number of further applications of this similarity index for information retrieval applications.

We foresee multiple research paths as natural follow-ups to the ideas proposed herein as well as potential applications. Framing malware characterization as a tagging task not only provides a common lightweight description framework that can be extended to various detection and analysis technologies, but also allows for malware clustering, and alerts prioritization. 

We are also interested in expanding the set of tags used to describe malware samples to a more, potentially hierarchical, taxonomy as well as experimenting with constrained embedding spaces such as hyperbolic spaces.

\section*{Acknowledgment}

We thank Adarsh Kyadige, Andrew Davis, Hillary Sanders, Joshua
Saxe, Richard Harang, and Younghoo Lee for their suggestions and
feedback during the development of this research. We also thank
Richard Cohen for sharing his expertise in malware detection. This
research was funded by Sophos Ltd.

\bibliographystyle{IEEEtran}

\newpage
\bibliography{references}

\newpage

\appendices
\section{Tag definitions}
\label{apendix:definitions}
\begin{itemize}
\item Downloader: Malicious program whose primary purpose and functionality is to download additional content. Often similar in usage to a Dropper.
\item Dropper: Malicious program that carries another program concealed inside itself, and \emph{drops} that program onto an infected machine.
\item Ransomware: Malware whose goal is to encrypt or otherwise make inaccessible a user's files, to then demand payment to regain access to them.
\item Crypto-miner: A program that uses a machine's computational resources to mine cryptocurrency, without the user's knowledge or consent, sending the results back to a central location.
\item Worm: Software that automatically spreads itself.
\item Adware: Potentially unwanted software that shows the user an excessive number of - often in browser - ads, or changes the user's home page to an ad, to get more clicks.
\item Spyware: Covers programs that collect confidential information and send it to an attacker. This confidential information could range from web browsing habits, keystroke logging, password stealing or banking information among others.
\item Flooder: Designed to overload a machine's network connections. Servers are common targets of these attacks. 
\item Packed: Indicates that the malware was packed for the sake of avoiding detection.
\item File-Infector: Infects executable files with the intent to cause permanent damage or make them unusable. A file-infecting virus overwrites code or inserts infected code into a executable file.
\item Installer: Installs other unwanted software.
\end{itemize}

\section{Labeling details}
\label{app:labeling}

\subsubsection{Token Extraction}
The first step for deriving tags from detection names is parsing the individual detection names to extract relevant \emph{tokens} within these names.
A token is defined as a sequence of characters in the detection name, delimited by punctuation, special characters or case transitions from lowercase to uppercase (we create tokens both splitting and not splitting on case transitions). These are then normalized to lowercase. For example, from the detection name \emph{Win32.PolyRansom.k} we extract the set of tokens \{\emph{win32}, \emph{polyransom}, \emph{poly}, \emph{ransom}, \emph{k}\}. Once all the tokens from all the vendor detection names for a given dataset are created, we keep those tokens that appear in a fraction of samples larger than $\alpha$ in our dataset. In practice we set the threshold $\alpha$ to 0.06\%, resulting in 1,500 unique tokens. A manual inspection of the tokens with lower occurrence rates found that they were mostly non-informative pseudo-random strings of characters usually present in detection names (e.g. `31e49711', `3004dbe01').

\subsubsection{Token to Tag Mapping}
\label{sec:tag_creation}
Once the most common tokens were defined, we manually built an association rule from tokens to tags for those tokens related with well-known malware family names or those that could be easily associated with one or more of our tags. For example, \emph{nabucur} is the family name of a type of ransomware and therefore can be associated with that tag. Similarly, the token \emph{xmrig}, even though it is not the name of a family of malware can be recognized as referring to a crypto-currency mining software and therefore can be associated with the \emph{crytpo-miner} tag.
This way, we created a mapping from tokens to tags based on prior knowledge. With this mapping, we can now associate a sample with a tag if any of the tokens that map to that tag are present in any of the detection names given by the set of trusted vendors.

\subsubsection{Token Relationship Mining}
\label{sec:token_rel_mining}
In order to understand how tokens relate to each other, we compute the empirical token conditional probability matrix $\mathbf{K}$, 
    \begin{align}
        \label{eq:token_cond_prob}
        \mathbf{K}(i, j) = \tilde{p}(k_i | k_j) = \# (k_i \cap k_j) / \# k_j,
    \end{align}
\noindent 
where $\#k_j$ is the number of times the token $k_j$ appears in a given dataset, and $\#(k_i \cap k_j)$ is the number of times $k_i$ and $k_j$ occur together. $\mathbf{K}(i, j)$ is then, by definition, the empirical conditional probability of token $i$ given token $j$ for a given dataset of samples.
We then define the following pairwise relationships between tokens based on their empirical conditional probabilities:
\begin{itemize}
    \item Tokens $k_i$ and $k_j$ are \emph{synonyms} under threshold $\beta$ if and only if
    $\tilde{p}(k_i|k_j) > \beta$ and $\tilde{p}(k_j|k_i) > \beta$.
    \item Token $k_i$ is a parent of token $k_j$ under threshold $\beta$ if and only if $\tilde{p}(k_i|k_j) > \beta$ and $\tilde{p}(k_j|k_i) \leq \beta$.
    \item Token $k_i$ is a child of $k_j$ under threshold $\beta$ if and only if $\tilde{p}(k_i|k_j) \leq \beta$ and $\tilde{p}(k_j|k_i) > \beta$.
\end{itemize}

With this in mind, we extend our previously proposed labeling function as follows. We use the tag $t_i$, associated with a set of tokens $K_i = \{k_1, \dots, k_l\}$, to describe a given malware sample $\mathbf{x}$ if, after parsing the detection names for $\mathbf{x}$ we find that:
    \begin{itemize}
        \item any of the tokens $k \in K_i$ is present for sample $\mathbf{x}$,
        \item OR any of the synonyms of $k$ is present for the sample (for every $k \in K_i$),
        \item OR any of the children of $k$ is present (for every $k \in K_i$).
    \end{itemize}
\noindent
The first bullet refers to the use of a manually created mapping between tags and tokens, i.e. our original labeling function. The following two bullets define automatic steps for extending the tag definitions and improving the stability of the tagging method. Empirically, we observed that when computing the token co-occurrence statistics in our training set as in Equation \ref{eq:token_cond_prob}, the automatic steps improved the tag coverage in the validation set in average by 13\%, while increasing the mean token redundancy, or the mean number of tokens observed per tag from 2.9 to 5.57.
This increase in mean token redundancy makes the labeling function more stable against mis-classifications or missing scans from the set of trusted vendors. 

The parameter $\beta$ was set to 0.97, the value at which the coverage of malicious tags improved for malware samples in our validation set, while remaining constant for benign samples. 
file read and writes operations, as well as many other activities can be captured that would not necessarily be observable in a static scan alone as in its packaged state this data would be encrypted, or possibly not even present as it may be downloaded at the time of execution.  Sophos researchers have created family specific signatures for this scanbox environment that are able to analyze dropped, downloaded and modified files, as well as memory dumps, network traffic, and many other artifacts in addition to static binany scans to develop dynamic signatures with much more stringent criteria to define family membership.

\begin{table*}[t]
\caption{Per-tag evaluation results for the two proposed architectures on all (malicious and benign) samples of the test dataset. Both recall and F-score are computed by binarizing each classifier’s outputs at a false positive rate of 0.01 on the entire set for each tag. The weighted mean weights the contribution of each tag by its support. The best result between the two proposed architectures for each row is highlighted in bold.}
\centering
\begin{tabular}{l|ccc|ccc}
\multicolumn{1}{c|}{} & \multicolumn{3}{|c|}{Multi-Head}  & \multicolumn{3}{c}{Joint Embedding} \\
Tag name        & AUC               & \begin{tabular}[c]{@{}c@{}}Recall \\ @FPR=$10^{-2}$ \end{tabular}   & \begin{tabular}[c]{@{}c@{}} F-score \\ @FPR=$10^{-2}$ \end{tabular}
                & AUC               & \begin{tabular}[c]{@{}c@{}}Recall \\ @FPR=$10^{-2}$ \end{tabular}   & \begin{tabular}[c]{@{}c@{}} F-score \\ @FPR=$10^{-2}$ \end{tabular}         \\ \hline \hline
adware          & \textbf{0.984}        & \textbf{0.847}        & \textbf{0.861} 
                & 0.983                 & 0.845                 & 0.860                 \\
crypto-miner    & \textbf{0.999}        & 0.991                 & \textbf{0.854} 
                & \textbf{0.999}        & \textbf{0.992}        & \textbf{0.854}        \\
downloader      & 0.984                 & 0.550                 & 0.693 
                & \textbf{0.985}        & \textbf{0.618}        & \textbf{0.747}        \\
dropper         & 0.977                 & 0.734                 & 0.835 
                & \textbf{0.978}        & \textbf{0.760}        & \textbf{0.851}        \\
file-infector   & 0.996                 & \textbf{0.968}        & \textbf{0.967} 
                & \textbf{0.997}        & \textbf{0.968}        & \textbf{0.967}        \\
flooder         & 0.985                 & \textbf{0.970}        & \textbf{0.524} 
                & \textbf{0.988}        & 0.969                 & 0.523                 \\
installer       & 0.994                 & 0.943                 & 0.813 
                & \textbf{0.995}        & \textbf{0.945}        & \textbf{0.814}        \\
packed          & 0.993                 & \textbf{0.914}        & \textbf{0.946} 
                & \textbf{0.994}        & 0.905                 & 0.941                 \\
ransomware      & \textbf{0.998}        & \textbf{0.987}        & \textbf{0.939} 
                & \textbf{0.998}        & 0.986                 & 0.938                 \\
spyware         & 0.980                 & 0.746                 & 0.847 
                & \textbf{0.982}        & \textbf{0.749}        & \textbf{0.849}        \\
worm            & 0.990                 & 0.898                 & 0.926 
                & \textbf{0.992}        & \textbf{0.905}        & \textbf{0.929}        \\ \hline
mean            & 0.989                 & 0.868                 & 0.837 
                & \textbf{0.990}        & \textbf{0.877}        & \textbf{0.843}       \\
weighted mean   & 0.987                 & 0.818                 & 0.872 
                & \textbf{0.988}        & \textbf{0.829}        & \textbf{0.881} 
\end{tabular}
\label{tab:pertag_results_s}
\end{table*}

\section{Evaluation in complete test dataset}
\label{sec:eval_with_benign}

Table \ref{tab:pertag_results_s} shows the results of the tagging models for a complete test set, which includes 328,458 benign samples on top of the only-malware evaluation dataset of size 3,456,288 as introduced in Section \ref{sec:data_description}. We see that there is no degradation in performance when adding benign samples to the evaluation. This is an indication that the model has learnt not to assign malicious tags to benign samples.

\section{Sample statistics for anchor experiment}

Table \ref{tab:mal_fams} summarizes the families, the number of examples per family and their associated tags for the anchor experiment described in Section \ref{sec:embedding_comparison}. The two different crypto-miner families refer to alternate approaches to perform crypto mining in a host machine without administrator knowledge.

\begin{table}[t!]
\caption{List of notable malware families identified in the test set by our behavioral engine, along with the list of associated tags for each of the families, and the number of samples in $D_{\text{test}}$.}
\centering{
\begin{tabular}{l|l|r}
Campaign Name           & \multicolumn{1}{c|}{Associated Tags} & \multicolumn{1}{c}{n. samples} \\ \hline \hline
fareit                & downloader, spyware		& 1,949  \\
crypto\_miner\_groupA & crypto-miner			& 300    \\
upatre                & downloader				& 217    \\
crypto\_miner\_groupD & crypto-miner			& 198    \\
bladabindi            & spyware					& 186    \\
nanocore              & spyware					& 127    \\
shellter              & file-infector			& 96     \\
remcos                & spyware					& 61     \\
emotet                & spyware					& 57     \\
darkcomet             & spyware					& 46     \\
recam                 & spyware					& 43     \\
formbook              & spyware					& 35     \\
cerber                & ransomware				& 33                                   
\end{tabular}}
\label{tab:mal_fams}
\end{table}

\end{document}